\documentclass[11pt,a4paper]{article}
\usepackage[hyperref]{acl2020}
\usepackage{times}
\usepackage{latexsym}

\usepackage{microtype}

\usepackage{graphicx}
\usepackage{algorithm}
\usepackage{algorithmic}
\usepackage{amsmath}
\usepackage{amsthm}
\usepackage{booktabs}
\usepackage{colortbl}
\usepackage{multirow}
\usepackage{color}
\usepackage{amsfonts}
\usepackage{bm}
\usepackage{CJKutf8}
\usepackage{array}

\usepackage{subcaption}
\usepackage{url}

\aclfinalcopy 
\def\aclpaperid{73} 

\title{Listener's Social Identity Matters in Personalised Response Generation}

\author{Guanyi Chen\textsuperscript{$\spadesuit$}, 
Yinhe Zheng\textsuperscript{$\clubsuit\heartsuit$}, \and
Yupei Du\textsuperscript{$\spadesuit$}\\
\textsuperscript{$\spadesuit$}Department of Information and Computing Sciences, Utrecht University\\
\textsuperscript{$\clubsuit$}Samsung Research China - Beijing (SRC-B)\\
\textsuperscript{$\heartsuit$}Department of Computer Science and Technology, Tsinghua University\\
\texttt{g.chen@uu.nl, yh.zheng@samsung.com, y.du@uu.nl}}

\date{}

\begin{document}
\maketitle
\begin{abstract}
Personalised response generation enables generating human-like responses by means of assigning the generator a social identity. However, pragmatics theory suggests that human beings adjust the way of speaking based on not only who they are but also whom they are talking to. In other words, when modelling personalised dialogues, it might be favourable if we also take the listener's social identity into consideration. To validate this idea, we use gender as a typical example of a social variable to investigate how the listener's identity influences the language used in Chinese dialogues on social media. Also, we build personalised generators. The experiment results demonstrate that the listener's identity indeed matters in the language use of responses and that the response generator can capture such differences in language use. More interestingly, by additionally modelling the listener's identity, the personalised response generator performs better in its own identity.
\end{abstract}

\begin{CJK}{UTF8}{gbsn}

\section{Introduction} \label{sec:intro}

Persona plays an important role in our daily communication since it affects the way we render our dialogues. Social variables, such as gender, age, place of birth or even wealth and social status, account for a large proportion in each individual's persona. Numerous previous studies have suggested that these variables strongly affect each speaker's word preference in dialogues. A growing body of works has been carried out to implicitly or explicitly model these variables in dialogues~\cite{Li2016_ACL, Qian2017Assigning, Kottur2017Exploring, P18-1205, DBLP:journals/corr/abs-1901-09672, zheng2019pre}.

Despite the reported success, most previous studies for personalised dialogue modelling consider only the persona of speakers. \footnote{For using the terminology consistently, we use ``speaker'' referring to the person who produces the response (who is also a personalised dialogue system heading to model) and ``listener'' referring to the one who utters the post.} Nevertheless, the pragmatics theory suggests that the speaking style will be adjusted not only by who the speaker is, but also whom the speaker is talking to \citep{wish1976perceived, hovy1987generating}. In the computational linguistics community, \citet{dinan2020multi} investigates this issue by measuring and mitigating gender bias in dialogue dataset utilising a gender classifier. From the aspect of personalised dialogue generation, \citet{P18-1205} and \citet{zheng2019pre} tried to attach the listener persona to the encoder of their generator, but interestingly, they obtained very different results, namely, the performance of \citet{P18-1205} went down while that of \citet{zheng2019pre} went up.

\begin{figure}
    \centering
    \includegraphics[scale=0.55]{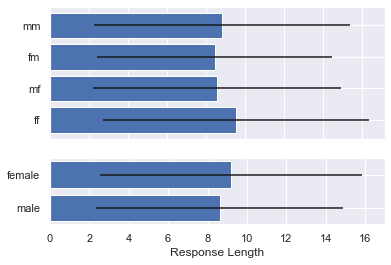}
    \caption{Average response length of each style.}
    \label{fig:langth}
\end{figure}

Nonetheless, no systematic studies have been conducted to investigate what role does the listener's identity play in personalised response generation. Research questions that we wish to answer by the proposal put forward in this paper are:
\begin{enumerate}
    \item How the listener's social identity impacts the responder's language use;
    \item Can a response generator capture this impact, if yes, in which way?
\end{enumerate}
To this end, we apply analysis and build a response generator on a Chinese personalised dialog dataset: \textsc{PersonalDialog}, a corpus extracted from Weibo\footnote{Weibo is the largest Chinese social media.}. There are two reasons to use this dataset: one is that the \textsc{PersonalDialog} dataset origins from the real conversations on social media Weibo, in which speakers' social variables play an important role; the other is that this dataset provides a massive amount of dialogue data (over 20M sessions) between a large population of speakers (over 8M speakers). It is of sufficient size to capture a variety of linguistic phenomena that are associated with social variables. Each speaker/listener in \textsc{PersonalDialog} comes up with 4 social variables: gender, age, location, and interests. For simplicity and for conducting controlled analysis and experiments, we only focus on gender in this paper.

As for the first research question, we postulate that a speaker behave differently when s/he speaks to people with different gender stylistically. This yields four possible speaking styles: \texttt{ff}, \texttt{mf}, \texttt{fm}, and \texttt{mm}\footnote{We use \texttt{fm} to represent the style used by a male speaker when talking to a female listener. Similar definition applies to \texttt{mf}, \texttt{mm}, and \texttt{ff}.}. We, therefore, build a classifier to separate these styles defining on ``gender-pairs''. Previous analysis on blogging data~\citep{schler2006effects, goswami2009stylometric, nguyen2011author,bamman2014gender} has identified that one of the key features for distinguishing contents produced by a female from those by a male is the sentence length, i.e., females tend to utter longer sentences. As shown in Figure~\ref{fig:langth}, the same phenomenon is found in \textsc{PersonalDialog}: females' responses are generally longer than males'. Further statistics on the response length falling the above four styles suggest that gender-pairs are also separable, perhaps excepting \texttt{mf} and \texttt{fm} at first glance. To validate this and understand why, we build a gender-pair classifier and conduct so-called pivot word analysis. We find out which word contributes the most for helping the classifier make decisions. Experiment results show that these styles are separable, but \texttt{mf} and \texttt{fm} are often confused with each other.

As for the second research question, we build a personalised response generator conditioning on these styles. The outcomes suggest that the generator could capture the difference between those styles and, in addition, modelling the listener's identity helps the generator to express its own identity. Moreover, based on previous analyses, we have also tried to merge the style of \texttt{mf} and \texttt{fm} into a single integrated style \texttt{mf/fm}. However, the final results of the response generator suggests that it is hard to model utterances with this integrated style.
\section{Gender-Pair Classification} \label{sec:classification}

To approach the first research question, we build a gender-pair classifier to simultaneously recognise the speaker's and listener's social identity based on the dialogue utterances. Concretely, as aforementioned in section~\ref{sec:intro}, we assume the present task as a style classification task and design four labels for each input dialogue utterance: \texttt{mm} (male talking to male), \texttt{mf} (female talking to male), \texttt{fm} (male talking to female), and \texttt{ff} (female talking to female).

However, in light of the Linguistic Style Matching theory~\citep{niederhoffer2002linguistic}, speakers will imitate the linguistic style of their conversation companion to pursue higher engagement. In other words, when two different gendered speakers communicate with each other, their speaking style may assimilate to each other as the conversation proceed. On the top of this observation, one may say that dissociating \texttt{fm} and \texttt{mf} is hard, and, therefore, it would be favourable if we merge \texttt{fm} and \texttt{mf} into a single category, namely \texttt{mf/fm}.

\subsection{Build Gender-Pair Classifiers} \label{sec:classifier}

\begin{table}[t]
\centering
\begin{tabular}{lccc}
\toprule
 Model & 2-way & 3-way & 4-way \\
 \midrule
fastText & 0.85 & 0.75 & 0.68 \\
TextCNN & 0.85 & 0.73 & 0.63 \\
LSTM & 0.85 & 0.75 & 0.63 \\\midrule
BOW Classifier & 0.85 & 0.74 & 0.64 \\
 \bottomrule
\end{tabular}
\caption{F1 score of the gender-pair classifiers.}
\label{tab:classifier}
\end{table}

Building on what has been discussed, to further get insight from conventional gender classification, we consider the following three classification tasks basing on three speaking style categorisation schemes: 1) two-way classification: classifying only speakers' gender, in which two labels are used: \texttt{male} and \texttt{female}; 2) three-way classification: classifying the conversational texts based on a merged labelling scheme, i.e., three labels are considered \texttt{mm}, \texttt{fm/mf}, and \texttt{ff}; and 3) four-way classification: the gender-pair classification which classifies the conversational texts into \texttt{mm}, \texttt{fm}, \texttt{mf}, and \texttt{ff}.

\subsubsection{Classification Models} 

We test a number of text classification algorithms, including fastText\footnote{The official implementation of fastText from Facebook is used: \url{https://github.com/facebookresearch/fastText}.} \citep{joulin2017bag}, TextCNN \citep{kim-2014-convolutional} and LSTM \citep{hochreiter1997long} (in which the hidden states of all the tokens are max pooled before being feed into the final Softmax layer). In order to conduct interpretable analysis, we train a Bag-of-Word (BOW) classifier: a logistic regression with only unigram features.

\subsubsection{Experimental Settings} 

Building on the fact that classifying the social variables based on the social media data is hard~\citep{nguyen2013old, nguyen2014gender}, and the exhibition of speakers' social identities is sparse in social media text~\citep{zheng2019pre}, we adopt the classification strategy used by \citet{DBLP:journals/corr/abs-1901-09672}. Specifically, each classifier input is a concatenations of $N$ randomly sampled responses with the same style. In this study, we use $N=20$. We train and test the classifiers on \textsc{PersonalDialog}, where the dataset has been divided into training and testing sets without overlapping. The training data are down-sampled to balance the corpus. 10\% of the training set is held out for tuning parameters, and the final models are trained on the whole training set. The classifiers are evaluated using F1 scores.

\begin{figure}[t]
    \centering
    \includegraphics[scale=0.55]{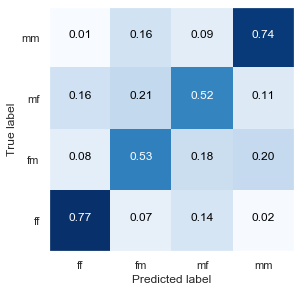}
    \caption{Confusion matrix of the 4-way gender-pair classification using fastText.}
    \label{fig:confusion}
    \vspace{-1em}
\end{figure}

\subsubsection{Experimental Results} 

Table~\ref{tab:classifier} depicts the performances of these classifiers. FastText performs remarkably well. It outperforms both TextCNN and LSTM, which are models having much higher complexity and capacity. It is surprising that the simplest BOW classifier also achieves comparably good performance, which suggests that the word usage is the most important feature for distinguishing speakers' social identity (at least for the gender). Further comparison of the fastText and BOW classifier embodies that the unigram features are sufficient for conducting gender classification in the coarse 2-way classification setting, while higher-ordered N-gram features (used by the fastText) are useful in more fine-grained 3-way and 4-way classification settings. 

The F1 score of the 4-way gender-pair classification using fastText reaches 0.68. 
This means that it is feasible to identify the style of the listener by only considering the utterances issued by the speaker.
We print the confusion matrix of this result in Figure~\ref{fig:confusion}.
The utterances from \texttt{ff} and \texttt{mm} are rarely confused with each other. 
This indicates that the language use of both males and females have clear differences when they speak to people with the same gender
When they talk to people with different gender, in line with the results of gender classification, they tend to express stylistic characteristics related to their own gender since confusions appear between \texttt{fm} and \texttt{mm} as well as between \texttt{mf} and \texttt{ff}.
Nonetheless, we also observe equally severe confusion between \texttt{fm} and \texttt{mf}, which approves that the linguistic style matching hypothesis plays a certain role when people expressing their social identities.

In addition, we also observe a certain level of confusion between \texttt{fm} and \texttt{ff} as well as between \texttt{mf} and \texttt{mm}.
This said, the classifier sometimes confuse between, for example, an utterance from a male and an utterance from a female when they both speak to male listeners. This, yet again, could be seen as an evidence for the existence of linguistic style matching.
Although the utterance from \texttt{fm} and \texttt{mf} shows a tendency of assimilation, it appears that the speakers still maintain the characteristics of their own gender and, in this sense, there are still certain reasons to disassociate the style of \texttt{fm} from \texttt{mf}.

\subsection{Pivot Word Discovery}

\begin{algorithm}[t]
    \caption{Classifier-based Pivot Word Discovery}
    \begin{algorithmic}[1]
    \REQUIRE Dataset $\mathcal{D}$, Style Set $\mathcal{S}$, BOW Classifier $f$, Confidence Threshold $\alpha$, and Word Pivot Frequency Threshold $\beta$.
    \ENSURE A set of Pivot Words $\Omega$
    \FOR{each input sentence $x$ and corresponding label $y=s \in \mathcal{S}$ in $\mathcal{D}$}
    \STATE Predict label $\hat{y}$ and confidence $p$ for $x$
    \IF{$\hat{y}$ = $y$}
    \FOR{each word type $t$ in $x$}
    \STATE Construct $x_{\backslash t}$ by removing all $t$ in $x$
    \STATE Predict label $\hat{y}'$ and confidence $p'$ for $x_{\backslash t}$ 
    \IF{$\hat{y}' \neq y$ or $p - p' > \beta$}
    \STATE Add $t$ to $\Omega_c$ and add pivot word frequency $p(t, s)$ by 1
    \ENDIF
    \ENDFOR
    \ENDIF
    \ENDFOR
    \RETURN All $t$ in $\Omega_c$ if $p(t, s) > \beta$ for all $s \in \mathcal{S}$
    \end{algorithmic}
    \label{alg:pwd}
\end{algorithm}

To understand how people change their language use with respect to social identities of themselves and of whom they speak to, or, in other words, to understand how the gender-pair classifiers make their decisions, we apply the \emph{Pivot Word Analysis}. Pivot words are words that have substantial influence on the classifier's decision making and have been widely used for interpreting the language use in many language generation tasks such as Style Transfer~\citep{fu-etal-2019-rethinking} and Table-to-Text Generation~\citep{ma-etal-2019-key}.


\subsubsection{Pivot Word Extraction Algorithm.} 

Since the expression of social identity is sparse in the social media data, the appearance of pivot words in the utterance is also sparse. Therefore, the pivot word discovery algorithms introduced in~\citep{fu-etal-2019-rethinking} and~\citep{ma-etal-2019-key} are not applicable in the present task. Instead, we use a simple yet efficient pivot word discovery algorithm coined as \emph{Classifier-based Pivot Word Discovery} for extracting pivot words using the trained BOW classifier. 

The algorithm is of finding out which word type in the training data plays a major role in the BOW classifier's decision-making. It is sketched in Algorithm~\ref{alg:pwd}. As can be seen from lines 2-5, this algorithm only considers samples that have been correctly classified. For each word type $t$ in a sample $x$, it compares the classification results and confidences when including and excluding $t$ in $x$ (lines 2-8). Specifically, if the classifier's predicted result is changed or the prediction confidence's change exceeds a certain threshold of $\beta$, we extract it as a pivot word candidate (line 10). If the same word type has been extracted as a candidate for more than $\alpha$ times under a single category, the algorithm returns it as a pivot word (line 15). In this work, we set $\alpha$ and $\beta$ to 10 and 0.5, respectively.

\begin{table*}[t]
\centering
\small
\begin{tabular}{lp{14cm}}
\toprule
 Model & Example Pivot Words \\
 \midrule
\texttt{mm} & 华为 (Huawei), 苹果 (Apple), 三星 (Samsung), 小米 (Xiaomi), 美国 (America), 日本 (Japan), 中国 (China), 大陆 (Mainland), 台湾 (Taiwan),``，'', ''。''  \\ \midrule
\texttt{fm} & 游戏 (game), 王者 (Honer of Kings), 早安 (good morning), 晚安 (good night), 拍照 (photograph), 读书 (reading), 工作 (working), 我 (I), 你 (you)\\ \midrule
\texttt{mf} & 大叔 (Uncle), 弟弟 (little Brother), 哥哥 (elder Brother), 上班 (Working), 喝酒 (Drinking), 厦门 (Xiamen), 广东 (Guangdong), 广州 (Guangzhou), 嗯嗯 (Uh-huh), 我 (I), 你 (you), ``$\sim\sim$'', ``!!!!'', ``???''\\ \midrule
\texttt{ff} & 王俊凯 (a celebrity), 易烊千玺 (a celebrity), 鹿晗 (a celebrity), KPop, 男主 (leading actor), 电视剧 (teleplay), 化妆 (make up), 漂亮 (beauty), 裙子 (skirt), 便宜 (cheap), 淘宝 (Taobao)， 嗯嗯 (Uh-huh), 啊啊啊 (Ah Ah Ah), 我 (I), 你 (you), ``$\sim\sim\sim\sim$'', "!!??", "!!!!"\\ \midrule\midrule
\texttt{male} & 华为 (Huawei), 苹果 (Apple), 美国 (America), 大陆 (Mainland), 台湾 (Taiwan), 妹子 (girl), 媳妇 (wife), 游戏 (game), ``，'', ''。'' \\ \midrule
\texttt{female} & 王俊凯 (a celebrity), 易烊千玺 (a celebrity), 男主 (leading actor), 电视剧 (teleplay), 化妆 (make up), 裙子 (skirt), 面膜 (mask), 刘海 (bang), 我 (I), ``$\sim\sim$'', ``!!!!'', ``$\sim\sim\sim\sim$'', hhh, QAQ, mua\\ 
 \bottomrule
\end{tabular}
\caption{Lists of extracted pivot words in each categories of the gender classifier and the gender-pair classifier.}
\label{tab:pwd}
\end{table*}
\begin{table*}[t]
\centering
\begin{tabular}{ccccccc}
\toprule
  & \texttt{mm} & \texttt{mf} & \texttt{fm} & \texttt{ff} & \texttt{male} & \texttt{female}\\
\cmidrule(lr){2-5} \cmidrule(lr){6-7}
\texttt{mm} & 0.07 (-0.70) & 0.96 (+0.19) & 0.99 (+0.22) & 0.98 (+0.21) & 0.12 (-0.65) & 0.99 (+0.22)\\
\texttt{mf} & 0.72 (+0.19) & \cellcolor{gray} 0.00 (-0.53) & 0.23 (-0.30) & 0.01 (-0.52) & 0.41 (-0.12) & \cellcolor{gray}0.00 (-0.53)\\
\texttt{fm} & 0.27 (-0.25) & 0.31 (-0.21) & 0.02 (-0.50) & 0.19 (-0.33) & 0.04 (-0.48) & 0.11 (-0.41)  \\
\texttt{ff} & 0.79 (+0.05) & 0.10 (-0.64) & 0.21 (-0.53) & \cellcolor{gray} 0.00 (-0.74) & 0.94 (+0.20) & \cellcolor{gray}0.00 (-0.74) \\
 \bottomrule
\end{tabular}
     \caption{Recall of two pivot free classification experiments. Labels in the first row indicates the source categories the labels in the first column are the target categories. In each cell, a ($\pm b$) means the recall is a and comparing to its original performance the score increases/decreases b.}
     \label{tab:pfc}
\end{table*}

\subsubsection{Extracted Pivot Words.} 

Table~\ref{tab:pwd} lists typical examples of the extracted pivot words in each category for the gender classifier and the gender-pair classifier. As for the gender classification, we observe that the general topics used by males and females have clear differences on Weibo. Specifically, males focus on the topic of digital products, politics, and games while females like talking about starstruck, teleplays, makeup, and shopping. It is worth noting that one reason that Weibo users concentrate on these topics is that most of them are young people according to the statistics in~\citet{DBLP:journals/corr/abs-1901-09672}. These topics might change if use data extracted in more recent years since the \textsc{PersonalDialog} dataset was crawled in 2018.

More interestingly, we also find that differences exist in the use of punctuation and pronouns. Males use punctuation in a more formal way on social media (in which comma and period are frequently used), but females eager to concatenate a sequence of punctuation to express certain emotions or speech acts (e.g., ``$\sim\sim$'', ``!!!!''). The first person pronoun was extracted as pivot word for the \texttt{female} category, which might suggest that males are more likely to drop pronoun on social media. \footnote{Chinese as a discourse based language, pro-drop~\citep{huang1984distribution} is much more common than that in, for example, English, especially when the dropped pronoun referring to one of the speakers in a conversation~\citep{chen-etal-2018-modelling}.} To say the last word on how the use of zero pronouns is affected by the speaker's social identity needs further research, which is not the focus of this paper.
 
As for comparing the extracted pivot words for the gender-pair classifier and the gender classifier, in line with the classification results detailed in section~\ref{sec:classifier}, we observe more overlaps between \texttt{female} and \texttt{ff} as well as \texttt{male} and \texttt{mm} than between \texttt{female} and \texttt{mf} as well as \texttt{male} and \texttt{fm}. When comparing the words from different gender-pair categories, we find that people would talk about different topics when they talk to people of the same gender and with a different gender. 
For example, when a female talks to another female, they discuss ``idols'' they like, shopping, and dressing, which are rarely mentioned when she talks to a male. These observations explain why utterances with style \texttt{mf} (\texttt{fm}) are separable from those with style \texttt{ff} (\texttt{mm}) and suggest that the identities of listeners really matter the way of how speakers speaking.
 
As for the linguistic matching hypothesis, some evidences have been found. For example, \texttt{fm} and \texttt{mf} shared some topics including travelling, studying, working or gaming. Moreover, first person pronouns are more likely to be used when males speaking to females, but similar matching not appears in the use of punctuation.

\subsection{Pivot Free Classification} \label{sec:pfc}

In order to quantify how the gender-pair influences the language use, we do a \emph{Pivot Free Classification} experiment, where the BOW classifier is evaluated on the test data, in which the pivot words from a certain category are removed. Since we care about, by removing the pivot words, how many samples of a category are mis-classified into other categories, we report the recall scores in Table~\ref{tab:pfc}. We test the performance of the gender-pair classifier ``attacked'' by pivot words extracted by the gender-pair and the gender classifier. We name the category on which we report the performance as the \emph{target category} and the category from which we extract the pivot words as the \emph{source category}.

On the basis of the results in Table~\ref{tab:pfc}, we have the following observations: \textbf{First}, the performance reduces to almost zero if the source and the target are the same categories, which implies that the extracted pivot words are those which actually bias the decision making of the classifier. \textbf{Second}, \texttt{ff} and \texttt{mm} are definitely separable as no impact is found when they ``attack'' each other. \textbf{Third}, in line with the previous findings and the linguistic style matching theory, \texttt{mf} and \texttt{fm} are highly confused with each other, which can be approved from two dimensions: 1) as source categories, they highly reduce each other's performance; 2) Pivot words from \texttt{female} have remarkably effects on not only \texttt{mf} and \texttt{ff} but also \texttt{fm}. \textbf{Fourth}, \texttt{mf} and \texttt{fm} are not exactly the same, since, for instance, the impact of \texttt{mf} on \texttt{ff} is clearly higher than that of \texttt{fm} on \texttt{ff}. \textbf{Last}, the style of a conversation for speakers with a different gender is more similar to the style of how females speak. 
\section{Personalised Response Generation}

For exploring the second research question, that is, can a personalised response generator capture the differences of language use when imitating a speaker talking to listeners with different social identities? We train multiple response generators conditioning on the three style categorising schemes mentioned in section~\ref{sec:classification}. We start by introducing the basic architecture of our generator and the experimental settings. We then describe the evaluation metrics we use, with which we evaluate and analyse the generators.

\subsection{The Personalised Response Generator}

\begin{figure*}[t!]
    \centering
    \includegraphics[scale=0.7]{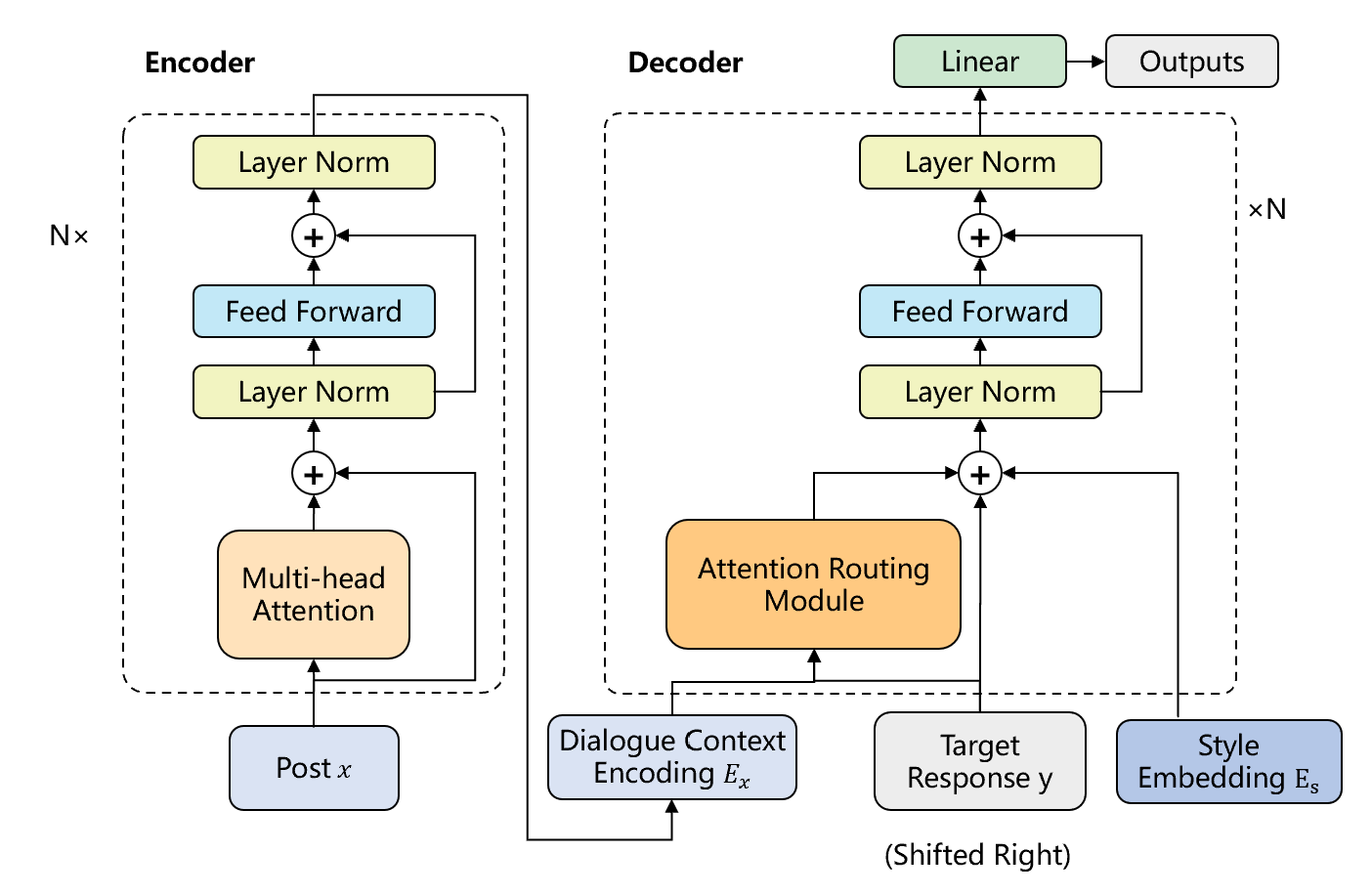}
    \caption{Illustration of our personalised response generator.}
    \label{fig:model}
\end{figure*}

Since inventing a new state-of-the-art personalised response generator falls out of the scope of this paper, we build the model following a simplified paradigm of~\citet{zheng2019pre,zheng2020stylized}. The architecture of the model we used is sketched in Figure~\ref{fig:model}. 

Concretely, given the dataset, containing $N$ dialogue pairs with each of their style: $\mathcal{D} = \{(x_1, y_1, s_1), ..., (x_N, y_N, s_N)\}$, where $x_i$ is the post, $y_i$ is the response, and $s_i$ is the style label of that response (i.e., in our case, it could be \texttt{female} or \texttt{fm}). As depicted in Figure~\ref{fig:model}, each post $x$ is firstly mapped into word embedding space using $\mathbf{e}_w(\cdot)$ and then is encoded via a Transformer~\citep{Vaswani2017Attention} based encoder to a representation $\mathbf{E}_x$.

\subsubsection{Encoding Style Information} 

Following~\citet{zheng2019pre}, in the decoding phase, we inject the style information by utilising the attention routing mechanism. Specifically, different from the standard Transformer decoder, both multi-head attention (MHA) and masked multi-head attention (MMHA) are deployed. In each decoder block, given the $\mathbf{E}_x$ and the embedded previously decoded response $\mathbf{E}_{y_{pre}} = \mathbf{e}_w(y_{pre})$, they are encoded to:
\begin{align}
    \mathbf{R}_{pre} & = \mbox{MMHA}(\mathbf{E}_{y_{pre}}, \mathbf{E}_{y_{pre}}, \mathbf{E}_{y_{pre}}) \\
    \mathbf{R}_{post} & = \mbox{MHA}(\mathbf{E}_{y_{pre}}, \mathbf{E}_x, \mathbf{E}_x)
\end{align}
Together with the mapped style, $\mathbf{E}_s$ is mapped using the style embedding $\mathbf{e}_s(s)$. These set of representations are merged in the following way to $\mathbf{R}$ before being feed for layer normalisation:
\begin{equation}
    \mathbf{R} = (\mathbf{R}_{pre} + \mathbf{R}_{post}) / 2 + \mathbf{E}_{y_{pre}} + \mathbf{E}_s
\end{equation}
in which, $\mathbf{R}_{pre}$ and $\mathbf{R}_{post}$ are averaged.

Despite of the simplicity, one major reason of why we do not use the original model of~\citet{zheng2019pre} in this study is that they did not encode personae (i.e., gender in our case) of speaker and listener symmetrically. To be more specific, they encode the persona of the listener as a number of style embeddings, which were added to the input together with the positional embeddings, while the speaker's persona was encoded as a sequence of words and was concatenated with the embeded post $x$. This kind of disassociation makes our experiments less controlled. Instead, in this study, we merge the label for speakers and listeners (i.e., the label such as \texttt{mf}) and map it into a single style embedding.

\subsubsection{Parameter Sharing and Pre-training.} 

Encoders and decoders in our model are sharing their parameters, which are initialised using a pre-trained Chinese GPT model~\citep{radford2019language,wang2020chinese}.

\subsection{Experiments} \label{sec:experiment}

\subsubsection{Experimental Settings}

We train and evaluate the model on the \textsc{PersonalDialog} dataset. For simplicity, in line with~\citet{DBLP:journals/corr/abs-1901-09672}, we only train and test our model using the first turn of each dialogue session in \textsc{PersonalDialog}. For conducting a controlled and fair analysis, we train three models corresponding to the three style categorisation schemes introduced in section~\ref{sec:classification} (see Table~\ref{tab:result}). In the following sections, we refer them with their ID, i.e., model 1, 2, or 3.

\subsubsection{Evaluation Metrics} 

Recall that our target is not of defeating state-of-the-art personalised response generator in the sense of generating better responses. Nonetheless, we still report some relevant results using commonly used automatic metrics including: BLUE~\citep{papineni2002bleu}, a metric comparing overlaps of n-grams ($n=1, 2$) between the reference responses and the generated responses for evaluating the adequacy and fluency; and DIST~\citep{li2015diversity}, measuring the proportion of distinct n-grams ($n=1$) for evaluating the diversity of the model outputs.

To help obtaining insights from the system outputs for the second research question, we design a number of new metrics based on the built classifiers and extracted pivot words from section~\ref{sec:classification}. Specifically, for evaluating a model with $n$ style categories, we propose the following metrics: 
\begin{enumerate}
\item\textbf{ACC.} evaluates whether the generated responses incorporate the target style using the trained $n$-way classifier. Similar approach is employed to evaluate the outputs of conditional language generators with off-line classifiers~\citep{zhou2017emotional, DBLP:journals/corr/abs-1901-09672, li-etal-2020-dual}. During evaluation, the system outputs are concatenated in the same way as the train data of those classifiers. Considering the speed and the performance, we use the fastText classifier in the evaluation;

\item\textbf{ACC-2.} evaluates whether the generated response reflect gender information using the trained gender classifier. It is worth noting that this metric is not applicable to model 2 since we have merged the \texttt{mf} and \texttt{fm}, we expect that they are no longer separable;

\item\textbf{Pivot Word Precision (PWP).} evaluates to what proportion the generated tokens are pivot words. Suppose the system outputs with style $s$ is $\hat{\mathcal{Y}}_s$  with the vocabulary $\mathcal{V}$ and the pivot words extracted by $n$-way classifier is $\Omega_s$, the PWP is computed by:
\begin{equation}
    \mbox{PWP}_s = \frac{\sum_{w \in \Omega_s} \#(w, \hat{\mathcal{Y}}_s)}{\sum_{w \in \mathcal{V}} \#(w, \hat{\mathcal{Y}}_s) }
\end{equation}
where $\#(w, \hat{\mathcal{Y}}_s)$ is the frequency of $w$ in $\hat{\mathcal{Y}}_s$. PWP is calculated for each style and is then micro-averaged;

\item\textbf{Pivot Word Recall (PWR).} evaluates how many word types in pivot words has been generated:
\begin{equation}
    \mbox{PWR}_s = \frac{\sum_{w \in \Omega_s} \mathbb{I}(w, \hat{\mathcal{Y}}_s)}{|\Omega_s|}
\end{equation}
where $\mathbb{I}(w, \hat{\mathcal{Y}}_s)$ equals to one if $w$ appears in $\hat{\mathcal{Y}}_s$, otherwise it equals to 0. 
\end{enumerate}

\begin{table*}[t]
\centering
\begin{tabular}{ccccccccc}
\toprule
 ID & Conditioned Styles & BLEU & DIST & ACC & ACC-2 & PWP & PWR\\
 \midrule
 1 & \texttt{female}, \texttt{male} & \textbf{3.94} & \textbf{0.092} & \textbf{76.47} & 76.47 & 41.44 & 66.04 \\
 2 & \texttt{ff}, \texttt{fm/mf}, \texttt{mm} & 3.58 & 0.089 & 48.00 & - & 46.93 & \textbf{69.15} \\
 3 & \texttt{ff}, \texttt{fm}, \texttt{mf}, \texttt{mm} & 3.60 & 0.089 & 63.03 & \textbf{84.00} & \textbf{55.37} & 63.05 \\
 \bottomrule
\end{tabular}
     \caption{Evaluation results of response generator with different speaking style categorisation scheme by means of metrics introduced in section~\ref{sec:experiment}.}
     \label{tab:result}
\end{table*}
\begin{table*}[t]
\centering
\begin{tabular}{ccccccc}
\toprule
  & \texttt{mm} & \texttt{mf} & \texttt{fm} & \texttt{ff} & \texttt{male} & \texttt{female}\\
\cmidrule(lr){2-5} \cmidrule(lr){6-7}
\texttt{mm} & 57.01 & 51.24 & 51.82 & 46.09  & 46.85 & 39.07\\
\texttt{mf} & 58.60 & 59.12 & 63.13 & 58.00  & 46.85 & 50.91\\
\texttt{fm} & 54.14 & 55.66 & 59.22 & 55.87  & 44.96 & 45.90\\
\texttt{ff} & 68.79 & 70.17 & 72.63 & 76.87  & 57.56 & 66.97\\
 \bottomrule
\end{tabular}
\caption{The results of cross-category PWR scores. Same as Table~\ref{tab:pfc}, categories in first row means where the pivot words from and categories in first column means where the system outputs from.}
     \label{tab:ccpwr}
\end{table*}

\subsubsection{Experimental Results}

Table~\ref{tab:result} charts the results of all the metrics above. It is not surprising that no significant difference is found in BLEU and DIST score between all three models since all of them have the same model architecture, the same parameter setting and, thus, the same capacity.

Due to the fact that different off-line classifiers have very different performance in their own domain (see Table~\ref{tab:classifier}), it is not fair to compare the value of ACC and ACC-2 across different dialogue generation models. However, taking other metrics into account, we still have some interesting findings. One is that all the ACC results are better than random, which somehow suggest that all of these models have captured the differences of language use under each style. The other is that although model 2 has the highest PWR and moderate level of PWP, but, meanwhile, it has the lowest ACC. In other words, it generates lots of pivot words, but the classifier does not classify them into the correct style. To understand why, we analysed the PWP for each style, and found that it works fine for \texttt{ff} (69.02) and \texttt{mm} (45.96), but collapses at the merged category, i.e., \texttt{mf/fm}. It obtains a PWP at only 25.82 and a PWR at 56.95 (which is not a very bad number). It appears that although the generator has produced fine amount of pivot words for expressing the style of \texttt{mf/fm}, but, the frequency of many of them might not be high. This also suggests that even though we found some evidences from experiments in section~\ref{sec:classification} supporting the theory of linguistic matching and the merging of \texttt{mf} and \texttt{fm}, but it seems that the generator we use cannot handle this.

More interestingly, we also find that model 3 not only has the highest performance on PWP, which means more than half of the tokens it produces are pivot words of the correct style, but also has the highest score on ACC-2 (i.e., the accuracy of gender classification), which is even better than model 1, a model that originally designed having two styles. This approves that by additionally modelling the social identities of the listeners, it helps the generator to utter more speaker identity related words because it takes the difference on speaking style when talking to listeners with different social identities into account.

\subsubsection{Cross-category PWR} 

To understand how model 3 works, we consider similar experiment to the one in section~\ref{sec:pfc} by measuring the cross-category PWR. From Table~\ref{tab:ccpwr}, we observe similar phenomenon as in section~\ref{sec:pfc}. For example, the pair \texttt{mm} and \texttt{ff} yields the lowest PWR when being as the pivot word source of each other. In contrast, they reach the highest score if they are their own pivot word source. \texttt{fm} and \texttt{mf} have relatively high PWR when being each other's pivot word source. When a male talks to an another male, they say very few words that females always say. Nevertheless, we also observe that sentences produced by \texttt{ff} always have the highest PWR regardless of where the pivot words are coming from. This should be a result of two reasons: most conversations in \textsc{PersonalDialog} dataset are between two females and PRR is a metric that sensitive to the size of test data (i.e., it is very likely that the more sentences are produced the more pivot words are included).
\section{Discussion}

We investigated the language use on Chinese social media regarding to the social identities of speakers and listeners. Specifically, we aim to explore whether the listener's social identities impact the responder's language use and whether such differences are separable.
The primary answers to both of these questions are ''Yes'' on the basis of our experiments and, additionally, by conducting pivot word analysis, we also found that \texttt{mf} and \texttt{fm} are less separable owing to the linguistic matching phenomenon. This raises as open question of which style categorisation scheme (i.e., whether to distinguishing \texttt{mf} and \texttt{fm} or not) is better for modelling personalised dialogues.

We then trained personalised response generators which take the social identities of listeners into account. To conduct insightful analysis, we design a number of new metrics with the help of the speaking style classifiers and the extracted pivot words. The outcomes show that modelling listener's identity assists the dialogue system to express more of its own identity. However, our system failed to model the style of \texttt{mf/fm}, which suggests the necessity of disassociating the style between \texttt{mf} and \texttt{fm}. 

Note that our work focus mainly on the gender, which from our perspective, underlies further studies on investigating the influence of other listener's social variables, such as age or location, or even of listener's persona as a whole. Likewise, since we study only on data from Chinese social media, it is also worth to validate whether our findings still hold in multilingual platforms like Twitter. As for the designing of dialogue systems, we highlighted the importance of modelling listener's persona for the Chatbot to express its own personality, it is also worthwhile to evaluate the built system in other angles, such as relevance and fluency, or to validate whether the resulting chat machine is empathetic~\citep{fung2018empathetic} or not.

Our decision on using single turn dialogue also limits the generalisability of our conclusion to real conversations since the assimilation of each others style may progress in the course of a dialogue. This may result in under-estimating the effect of the linguistic matching between speakers and listeners. In future, we will extend our work into multi-turn dialogue modelling. 

\section{Ethical Statement}

In this paper, we use the gender as an example of social identity to understand how the speaking style of a speaker is influenced. To this end, we build gender classifiers and stylised dialogue systems. In light of the discussion in~\citet{larson-2017-gender}, gender is notoriously difficult to detect~\cite{pmlr-v81-buolamwini18a}, and mis-gendering individuals is harmful to users~\cite{10.1145/3274357}.
Therefore, we are not and will not apply or extend the built classifiers and dialogue systems into real applications. We hope our findings could help with further works on mitigating gender bias~\citep{liu2020mitigating} or improving fairness~\citep{liu2019does} in dialogue systems.

\section*{Acknowledgements}
We thank the anonymous reviewers for their helpful comments.
Guanyi Chen is supported by China Scholarship Council (No.201907720022).

\bibliography{acl2020}
\bibliographystyle{acl_natbib}

\end{CJK}

\end{document}